\documentclass[letterpaper, 10 pt, conference]{ieeeconf}  
\IEEEoverridecommandlockouts                              
\usepackage{cite}

\usepackage{amsmath,amssymb,amsfonts}
\usepackage{algorithmic}
\usepackage{graphicx}
\usepackage{textcomp}
\usepackage{xcolor}
\usepackage{color}													
\usepackage{xfrac}														
\usepackage{colortbl}
\usepackage{siunitx}
\usepackage[english]{babel}	
\usepackage{multirow}
\usepackage{hyperref}
\usepackage{blindtext}
\usepackage{mathtools}
\usepackage{hyperref}
\usepackage[caption=false, font=footnotesize]{subfig}
\usepackage{tablefootnote}

\definecolor{olivengruen}{RGB}{110, 117, 14}

\newcommand{\Res}[0]{\boldsymbol{\mathit{\delta}}}

\title{\LARGE \bf
Towards Human-Robot Collaboration with Parallel Robots by Kinetostatic Analysis, Impedance Control and Contact Detection
}

\author{Aran Mohammad, Moritz Schappler and Tobias Ortmaier
	\thanks{All authors are with the Leibniz University Hannover, Institute of Mechatronic Systems, 30823 Garbsen, Germany,
		{\tt\small aran.mohammad@imes.uni-hannover.de}}%
}

\newif\ifcopyright
		\copyrighttrue

\begin{document}

	\ifcopyright
	{\LARGE IEEE Copyright Notice}
	\newline
	\fboxrule=0.4pt \fboxsep=3pt
	
	\fbox{\begin{minipage}{1.1\linewidth}  
			The final version of record is available at https://doi.org/10.1109/ICRA48891.2023.10161217\\   
			Copyright (c) 2023 IEEE. Personal use of this material is permitted. For any other purposes, permission must be obtained from the IEEE by emailing pubs-permissions@ieee.org. \\
			
			
	\end{minipage}}
	\else
	\fi

	\graphicspath{{./graphics/}}
	\maketitle
	\thispagestyle{empty}
	\pagestyle{empty}
	\begin{abstract}
		Parallel robots provide the potential to be leveraged for human-robot collaboration (HRC) due to low collision energies even at high speeds resulting from their reduced moving masses.
		However, the risk of unintended contact with the leg chains increases compared to the structure of serial robots. 
		As a first step towards HRC, contact cases on the whole parallel robot structure are investigated and a disturbance observer based on generalized momenta and measurements of motor current is applied. 
		In addition, a Kalman filter and a second-order sliding-mode observer based on generalized momenta are compared in terms of error and detection time. Gearless direct drives with low friction improve external force estimation and enable low impedance. 
		The experimental validation is performed with two force-torque sensors and a kinetostatic model. This allows a new identification method of the motor torque constant of an assembled parallel robot to estimate external forces from the motor current and via a dynamics model. 
		A Cartesian impedance control scheme for compliant robot-environmental dynamics with stiffness from \SI{0.1}{}--\SI{2}{\newton / \milli \meter} and the force observation for low forces over the entire structure are validated.
		The observers are used for collisions and clamping at velocities of \SI{0.4}{}--\SI{0.9}{\meter / \second} for detection within \SI{9}{}--\SI{58}{\milli \second} and a reaction in the form of a zero-g mode. 
	\end{abstract}

	\section{Introduction}
		Human-robot collaboration (HRC) is a current field of research that already lead to many industrial applications and commercial products for serial collaborative robots (cobots). 
		To enable safety regarding force and energy limits, serial cobots are programmed to a low speed and therefore suffer from increased cycle times in some industrial applications.
		Unlike their serial counterparts, a parallel robot (PR) consists of several parallel kinematic chains closing at a mobile platform \cite{Merlet.2006}. 
		PRs are characterized by drives mounted typically fixed to the robot base reducing the moving mass of each kinematic chain and allowing higher speeds while maintaining the same energy thresholds regarding HRC.
		\subsection{Related Work}
\begin{figure}[th!]
				\centering
				\includegraphics[width=1\columnwidth]{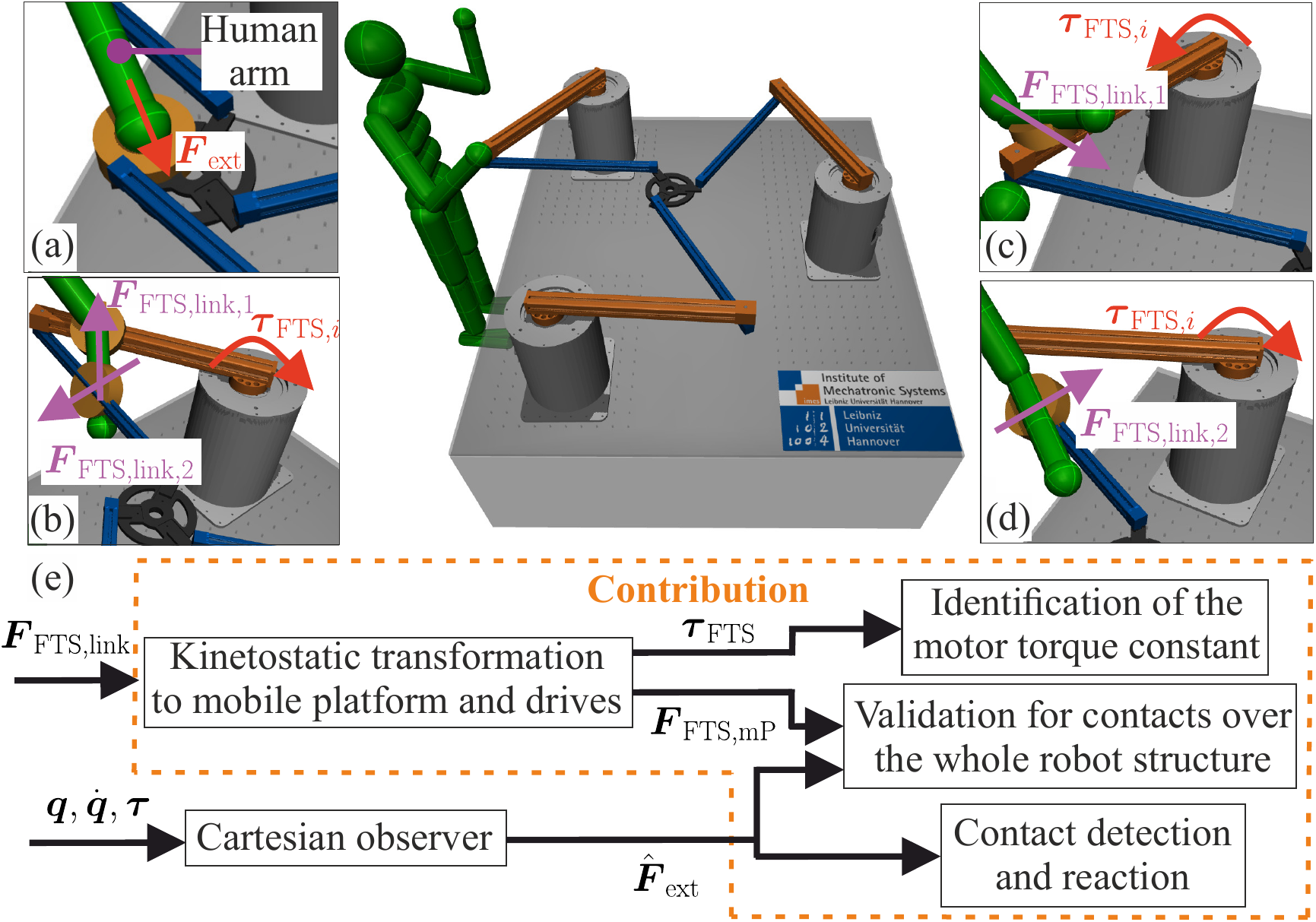}
				\caption{Unintended contact cases presented in a MuJoCo \cite{Todorov.2012} simulation: (a) platform collision, (b) link clamping, (c) and (d) link collision. In (e) is the contribution of this work presented: a projection of contact forces $\boldsymbol{F}_\mathrm{FTS,link}$ measured by a force-torque sensor (FTS) to the platform and (actuated) joint coordinate spaces is designed to identify the motor torque constant, validate the disturbance observer for contacts on the robot structure and evaluate the contact detection for collisions and clamping}
				\label{fig_collision_clamping}
			\end{figure}

			An ongoing topic in HRC is \emph{handling contacts} between robot and human. 
			Unintended contacts can be distinct in terms of their duration of impact and the possibility of withdrawal \cite{InternationalOrganizationforStandardization.2016}. 
			Figure~\ref{fig_collision_clamping}(a)--(d) shows possible unintended contact cases, which can be divided into clamping or collisions. 
			In the design phase, \emph{clamping risks} are reduced by rounding edges or avoiding small distances between the outer shells of the robot links \cite{Kock.2011}. 
			Light-weight robot links decrease \emph{collision-related risks} by reducing the kinetic energy.
			Tactile skin is able to detect contacts between humans and robots \cite{Dahiya.2013} together with data-driven models to classify intended and unintended contacts \cite{Albini.2017}. 
			To counter the need to integrate additional sensors into the robotic system, sensors already built in the robot can also be used for detection. 
			Physically motivated algorithms can be formed for \emph{contact detection} and \emph{isolation} (localization), as well as force \emph{identification}, by monitoring quantities through a residual, such as energy or generalized momentum \cite{Luca.2003, Luca.2006, Haddadin.2017}. 
			In \cite{Vorndamme.2021}, an observer of generalized momentum based on \emph{proprioceptive} sensing of the humanoid robot Atlas is analyzed and tested in a simulative study to perform real-time estimation of multiple contact forces and locations from distal to proximal links. 
			In \cite{Laha.2021b}, a circular-field coupling approach using a set-based task-priority strategy is followed to operate a two-arm system acting as one closed kinematic chain along a collision-free trajectory for wrench estimation and admittance-controlled responses to intended, as well as unintended contacts. 	
			In \cite{Wahrburg.2015}, the generalized-momentum observer is integrated into a Kalman filter 
			to parameterize modeling inaccuracies and measurement noise by the covariance matrices to estimate contact forces at the TCP of the dual-arm cobot ABB YuMi. An observer with finite-time behavior is presented in \cite{Garofalo.2019} for wrench estimation and collision detection on a KUKA LWR 4. The finite-time behavior is achieved by extending the momentum observer with a sliding-mode approach with a linear component. 
		
			The steps of contact detection to reaction are also necessary for the use of \emph{PRs in HRC}. Other than collaborative applications, compliant control or external force estimation of a PR were investigated in \emph{rehabilitation} \cite{Tsoi.2008, Tsoi.2009, Meng.2014, Jamwal.2016, Saglia.2013, Cazalilla.2016}, as a \emph{haptic device} \cite{Ergin.2011, Mitsantisuk.2015}, as academic \emph{demonstrators} \cite{Vinoth.2014, Singh.2015, Harada.2010} or in \emph{industrial applications} \cite{Bruzzone.2003, Cheng.2018, Harada.2016, Latifinavid.2018}. 
			Table \ref{tab_soa} gives an overview of selected works. 
			\begin{table}[b!]
				\caption[State of the art on PRs with regard to model-based estimation of external force, control law, application area in the columns]{State of the art on PRs regarding HRC\tablefootnote{Abbreviations are Impedance (I), Admittance (A), Force (F), Sliding Mode (SM) control, Disturbance Observer (DO), Extended Kalman Filter (EKF), Inverse Dynamics (ID)}} 
				\label{tab_soa}
				\begin{center}
					\begin{tabular}{|c|c|c|c|}
						\hline
						Ref.&Ext. Force estimated&Control&Application\\
						\hline
						\cite{Tsoi.2008}			&-&I&Reha\\			
						\cite{Tsoi.2009}			&Joint DO&F&Reha\\
						\cite{Meng.2014}			&-&I, A&Reha\\
						\cite{Jamwal.2016}		&-&I \& (SM)&Reha\\
						\cite{Saglia.2013}		&-&A&Reha\\
						\cite{Cazalilla.2016}	&-&A&Reha\\
						\hline
						\cite{Ergin.2011}		&-&I&Haptic Dev\\
						\cite{Mitsantisuk.2015}	&Joint DO&F&Haptic Dev\\
						\hline
						\cite{Vinoth.2014}		&EKF&I&Demo\\
						\cite{Singh.2015}		&Velocity DO&SM&Demo\\
						\cite{Harada.2010}		&-&I&Demo\\
						\cite{Dutta.2019}		&ID&I&Demo\\
						\hline
						\cite{Bruzzone.2003}		&-&I&Assembly\\
						\cite{Cheng.2018}		&motor current (static)&Position&Pick \& Place\\
						\cite{Harada.2016}		&-&I&Polish\\			
						\cite{Latifinavid.2018}	&-&A&Grinding\\
						\hline	
						\cite{Metillon.2022}	&-&A&HRC \tablefootnote{Out of scope of this work since it is a cable-driven PR}\\
						Ours				&Momentum-Based DO&I&HRC\\			
						\hline	
					\end{tabular}
				\end{center}	
			\end{table}
			In \cite{Meng.2014}, a Stewart platform is investigated for a 3T3R rehabilitation task, adapting the damping of the impedance to the user's electromyography signals in a data-based manner. 
			A modified Delta robot is designed in \cite{Ergin.2011} as a haptic interface with respect to different performance criteria in the workspace and operated with impedance control.
			\emph{Direct drives} are used so that the contact forces are visible in the drive's torque and are not influenced by gear friction to allow lower control impedances to be achieved. 
			In addition, contact detection via the motor current is favored. 
			For PRs in sensitive HRC, the use of torque sensors on the gear output side can thus be omitted. 
			Some \emph{industrial applications} require interaction with defined impedance characteristics.
			This is e.g. demonstrated by \cite{Bruzzone.2003} using a direct drive with a slider crank or by \cite{Harada.2016} using a slot-less direct drive motor.
		\subsection{Contributions}
			In the discussed literature, the use of PRs for interaction with humans happens only at a defined location on the mobile platform. 
			For the described use cases, the handling of unintended contacts over the entire \emph{robot structure} is not necessary, but for an HRC with PRs \emph{collisions} and especially \emph{clamping} at the parallel leg chains have to be considered. 	
			If detection, isolation and identification are based on the motor current, knowing the motor torque constant is decisive. 
			Based on possibly inaccurate data sheets, too high or low estimates lead to an incorrect determination of the contact force. 
			Identifying the motor torque constant either requires a time-consuming disassembly of the PR or a combined identification with the dynamics parameters, which may be error-prone. 
			An alternative by using a kinetostatic transformation of the contact forces on the drives will be introduced in this work, which is depicted in Fig.~\ref{fig_collision_clamping}(e). 
			Thus, the motor torque constant of an already assembled PR can be calibrated with measured contact forces on the robot structure and the motor current measurement.
			To ensure \emph{compliance over the entire robot structure} in the contact case, impedance control is preferred over admittance control. 
			One distinction from serial robots is that a PR is affected by \emph{kinematic constraints} and the resulting \emph{constraint forces}. 
			To estimate external forces separately, the closed-loop kinematics will be considered in the dynamics modeling for disturbance observation. 
			To exploit the advantages of PRs also for HRC, methods from serial cobots must be transferred to PRs, where methods for the mentioned differences and challenges are not yet sufficiently explored. 
			The first steps towards this goal form the contributions of our work\footnote{Supporting video: \url{https://youtu.be/HaazrQsKVhY}} which are also shown in Fig.~\ref{fig_collision_clamping}(e):
			\begin{itemize}		
				\item A kinematic and kinetostatic model transforms contact forces of any contact point of a PR to the mobile platform and to the actuators to determine the motor torque constant and for validation of observers.	
				\item The robot reacts sensitively with a low-impedance-controlled response in case of an unintended contact at the robot structure, which is detected by a generalized momentum-based disturbance observer for PRs allowing a robot stop. 
				\item A comparison is made with a Kalman filter and a second-order sliding-mode observer based on the generalized momenta in terms of error and detection time.
				\item The disturbance observation and reaction to contacts consisting of collisions and clamping along the whole robot structure are experimentally validated for a PR.
			\end{itemize}
			The paper is structured as follows. Starting with kinematics and dynamics modeling in section \ref{section_preliminaries}, HRC methods like disturbance observer and impedance controller for PRs are presented. 
			In section \ref{section_validation}, the PR used in this work is described, followed by an evaluation of collisions on the platform and robot links. Section \ref{section_conlusions} concludes the paper. 
	\section{Preliminaries} \label{section_preliminaries}
		In this section, the basics of kinematics (\ref{sec:kinematics}) and dynamics modeling (\ref{sec:dynamics}) of the PR in this work are described. 
		Subsequently, the kinematic analysis of an arbitrary contact location is performed. 
		The Cartesian impedance control (\ref{sec:controller}) and the disturbance observers (\ref{sec:observer}, \ref{sec:observer_kf}, \ref{sec:observer_sosml}) are introduced. 
		\begin{figure}[htb!]
			\vspace{1.5mm}
			\centering
			\includegraphics[width=\columnwidth]{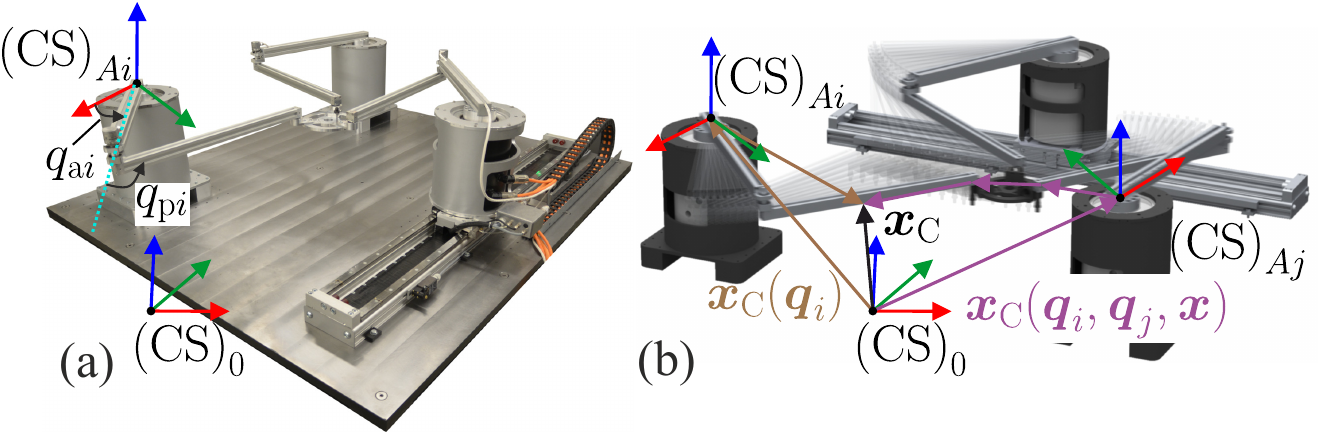}
			\caption{The 3-\underline{R}RR PR (a) with a contact at $_{(0)}\boldsymbol{x}_\mathrm{C}$ at the $i$-th leg chain (b) --- $_{(0)}\boldsymbol{x}_\mathrm{C}$ can be related to any leg chain and the joint angles}
			\label{fig_3PRRR_real_skizze}
		\end{figure}
		\subsection{Kinematics}
			\label{sec:kinematics}
			The methods presented in the following are generally applicable to any fully-parallel robot.
			The kinematics model is developed at the example of the planar 3-\underline{R}RR PR\footnotemark shown in Fig.~\ref{fig_3PRRR_real_skizze}(a) with $m{=}3$ platform degrees of freedom (DoF) and $n{=}3$ leg chains \cite{Thanh.2012}.\footnotetext{The letter R denotes a revolute joint and underlining actuation.}
			The PR has an actuated prismatic joint, which is kept constant in this work so that the position is not taken into account in the modeling. Operational space coordinates (platform pose), active, passive and coupling joint angles are represented respectively by $\boldsymbol{x}^\mathrm{T}{=}[r_x,r_y,\varphi_z]$, $\boldsymbol{q}_\mathrm{a} {\in}\mathbb{R}^n$ and $\boldsymbol{q}_\mathrm{p}, \boldsymbol{q}_\mathrm{c}$ with $\mathrm{dim}(\boldsymbol{q}_\mathrm{p}){=}\mathrm{dim}(\boldsymbol{q}_\mathrm{c}){=}3$.
			The $n_i{=}3$ joint angles (active, passive, platform coupling) of each leg chain in $\boldsymbol{q}_i{\in}\mathbb{R}^{n_i}$ can be represented by $\boldsymbol{q}^\mathrm{T}{=}[\boldsymbol{q}_1^\mathrm{T}, \boldsymbol{q}_2^\mathrm{T}, \boldsymbol{q}_3^\mathrm{T}] {\in} \mathbb{R}^{3n}$.
			\subsubsection{Inverse Kinematics}
				The inverse kinematics can be obtained by constructing the kinematic constraints $\Res (\boldsymbol{q}, \boldsymbol{x}){=}\boldsymbol{0}$ by means of vector loops \cite{Merlet.2006} and by eliminating the passive joint angles $\boldsymbol{q}_\mathrm{p}$ into the reduced kinematic constraints $\Res_\mathrm{red}(\boldsymbol{q}_\mathrm{a}, \boldsymbol{x}){=}\boldsymbol{0}$, yielding the explicit analytic formulation of the active joint angles in the form $\boldsymbol{q}_\mathrm{a} {=} \boldsymbol{\mathrm{IK}}(\boldsymbol{x, \boldsymbol{\sigma}})$. 
				The vector $\boldsymbol{\sigma}$ with $\mathrm{dim}(\boldsymbol{\sigma}){=}n$ contains a binary sign information of the elbow angle configuration of each serial kinematic chain to describe the ambiguity of $\boldsymbol{q}_\mathrm{p}$ due to working modes. 
			\subsubsection{Differential Kinematics}
				A time derivative of the kinematic constraints $\Res(\boldsymbol{q}, \boldsymbol{x})$ and $\Res_\mathrm{red}(\boldsymbol{q}_\mathrm{a}, \boldsymbol{x})$ leads to the linear transformations
				\begin{align}\label{eq_DifKin_Jac1}
					\dot{\boldsymbol{q}}&=-\Res_{\partial \boldsymbol{q}}^{-1}\Res_{\partial \boldsymbol{x}} 		\dot{\boldsymbol{x}}=\boldsymbol{J}_{q,x}\dot{\boldsymbol{x}}\\
					\label{eq_DifKin_Jac2}
					\dot{\boldsymbol{x}}&= -\left(\Res_\mathrm{red}\right)_{\partial \boldsymbol{x}}^{-1} \left(\Res_\mathrm{red}\right)_{\partial \boldsymbol{q}_\mathrm{a}} 		\dot{\boldsymbol{q}}_\mathrm{a}=\boldsymbol{J}_{x,q_\mathrm{a}}\dot{\boldsymbol{q}}_\mathrm{a}
				\end{align}
				with the notation $\boldsymbol{a}_{\partial \boldsymbol{b}}{\coloneqq} \sfrac{\partial \boldsymbol{a}}{\partial \boldsymbol{b}}$ and the Jacobian matrices $\boldsymbol{J}_{q, x}{\in}\mathbb{R}^{\dim(\boldsymbol{q})\times m}$ and $\boldsymbol{J}_{x, q_\mathrm{a}}{\in}\mathbb{R}^{m\times n}$. For the sake of readability, dependence on joint configuration $\boldsymbol{q}$ is omitted in the following.
			\subsubsection{Forward Kinematics}
				Forward kinematics requires more complex approaches, since platform poses of several working modes can have the same active joint angles \cite{Merlet.2006}. 
				To solve this problem, the passive joint angles $\boldsymbol{q}_\mathrm{p}$ are measured to calculate an initial estimate $\boldsymbol{x}_0$ of the platform pose together with the measured active joint angles. 
				Due to the different resolutions of the encoders, the Newton-Raphson approach is then used to calculate 
				\begin{align} \label{eq_FK_NR}
					\boldsymbol{x}_{i+1} = \boldsymbol{\mathrm{FK}}(\boldsymbol{q}_\mathrm{a}, \boldsymbol{q}_\mathrm{p}) = 		\boldsymbol{x}_i + \boldsymbol{J}_{x,q_\mathrm{a}} (\boldsymbol{q}_\mathrm{a}- \boldsymbol{\mathrm{IK}}(\boldsymbol{x}_i, \boldsymbol{\sigma}))
				\end{align} 
				until a predefined threshold is reached. The entries in $\boldsymbol{\sigma}$ can be determined by measuring the passive joint angles, by numerical integration of the joint velocities and plausibility analysis or by restricting the workspace.
			\subsubsection{Kinematic Analysis of an Arbitrary Contact Location}
				The previous considerations are limited to the mobile platform, which will now be extended to an arbitrary (contact) location on the robot structure. The starting point for this is the question \textit{how to represent the contact coordinates $_{(0)}\boldsymbol{x}_\mathrm{C}$ of a point $C$ via the joint angles $\boldsymbol{q}$}. 
				For this purpose, one assumes a contact at the $i$-th leg chain as visualized in Fig.~\ref{fig_3PRRR_real_skizze}(b), one first proceeds from a contact at the $i$-th leg chain, which is constant in the body-fixed coordinate system of the affected link. A virtual coordinate system is assumed at $\mathrm{C}$ and its pose is denoted by $_{(0)}\boldsymbol{x}_\mathrm{C}$ with the orientation of the body-fixed coordinate system. 
				The serial forward kinematics of the affected chain $i$ yields 	$_{(0)}\boldsymbol{x}_\mathrm{C}{=}\boldsymbol{f}_i( \boldsymbol{q}_i)$.
				At the same time, it is also possible to determine the searched pose via a different chain $j$ and the mobile platform in the form of $_{(0)}\boldsymbol{x}_\mathrm{C}{=}\boldsymbol{f}_j(\boldsymbol{q}_i, \boldsymbol{q}_j, \boldsymbol{x})$.
				Via the full kinematic constraints $\Res$ (including rotation) \cite{Schappler.2019}, 
				the platform orientation can be substituted resulting in $_{(0)}\boldsymbol{x}_\mathrm{C}(\boldsymbol{q}_i, \boldsymbol{q}_j)$.
				A differentiation w.r.t. time leads to $_{(0)}\dot{\boldsymbol{x}}_\mathrm{C}{=}\boldsymbol{J}_{x_\mathrm{C},q}\dot{\boldsymbol{q}}$, which contains the Jacobian matrix $\boldsymbol{J}_{x_\mathrm{C},q}$ as a linear transformation from the joint velocities to this contact's linear and angular velocity. 
				By (\ref{eq_DifKin_Jac1}) and (\ref{eq_DifKin_Jac2}) it follows that
				\begin{align}\label{eq_contactJacobian}
					_{(0)}\dot{\boldsymbol{x}}_\mathrm{C} &= \boldsymbol{J}_{x_\mathrm{C},q} \dot{ \boldsymbol{q}}\\
					&=\boldsymbol{J}_{x_\mathrm{C},q} \boldsymbol{J}_{q,x} \dot{\boldsymbol{x}} = \boldsymbol{J}_{x_\mathrm{C}, x} 		 \dot{\boldsymbol{x}} \\ 
					&= \boldsymbol{J}_{x_\mathrm{C}, x} \boldsymbol{J}_{x, q_\mathrm{a}} \dot{\boldsymbol{q}}_\mathrm{a} = 			\boldsymbol{J}_{x_\mathrm{C}, q_\mathrm{a}} \dot{\boldsymbol{q}}_\mathrm{a}.
				\end{align}
				By using the Jacobian matrices $\boldsymbol{J}_{x_\mathrm{C}, x}$ and $\boldsymbol{J}_{x_\mathrm{C}, q_\mathrm{a}}$, the kinematic relationship between the velocities is now identified, which will be used in the further analysis of the dynamics.
		\subsection{Dynamics} \label{sec:dynamics}
			The approach to dynamics modeling in this work is based on \cite{Thanh.2009} and provides for the derivation of the equations of motion in operational space based on \textsc{Lagrange}'s equations of the second kind, the \textit{subsystem} and \textit{coordinate partitioning} method to eliminate the kinematic constraint forces. 
			The inverse dynamics equation 
			\begin{equation} \label{eq_dyn}
				\boldsymbol{M}_x \ddot{\boldsymbol{x}}+ \boldsymbol{c}_x + \boldsymbol{g}_x+ \boldsymbol{F}_{\mathrm{fr},x}= 	\boldsymbol{F}_{\mathrm{m}} + \boldsymbol{F}_\mathrm{ext}
			\end{equation}
			applies for the present PR with the following generalized forces $\boldsymbol{F}{\in}\mathbb{R}^m$ (also including moments) acting on the platform. Equation \ref{eq_dyn} consists of $\boldsymbol{M}_x$ as the symmetric positive-definite inertia matrix, $\boldsymbol{c}_x{=}\boldsymbol{C}_x\dot{\boldsymbol{x}}$ as the vector/matrix of the centrifugal and Coriolis terms, $\boldsymbol{g}_x$ as the gravitational components, $\boldsymbol{F}_{\mathrm{fr},x}$ as the friction components consisting of viscous and Coulomb friction, $\boldsymbol{F}_{\mathrm{m}}$ as the generalized forces based on the motor torques and $\boldsymbol{F}_{\mathrm{ext}}$ as generalized external forces. In this work, the base dynamics parameters from \cite{Thanh.2012} are used. 
			In the identified values, the same geometrical and thus dynamics properties of the three leg chains are assumed, whereby the friction parameters are identified individually. 
			A transformation of the forces to the motor torques is done by the principle of virtual work $\boldsymbol{\tau}{=}\boldsymbol{J}_{x,q_\mathrm{a}}^\mathrm{T}\boldsymbol{F}$, which shows that a force on the platform has a configuration-dependent effect on all actuators and vice versa.
			Similarly, the effect caused by an external force $\boldsymbol{F}_\mathrm{ext,link}$ at a link can be transformed via the equations
			\begin{subequations}\label{eq_trafo_link_mP_Drives}\begin{align} 
				\boldsymbol{F}_\mathrm{ext}&=\boldsymbol{J}_{x_\mathrm{C},x}^\mathrm{T} \boldsymbol{F}_\mathrm{ext,link},\\
				\boldsymbol{\tau}_\mathrm{ext}&=\boldsymbol{J}_{x_\mathrm{C},q_\mathrm{a}}^\mathrm{T} \boldsymbol{F}_\mathrm{ext,link}.
			\end{align} \end{subequations}
		\subsection{Cartesian Impedance Control in Operational Space} \label{sec:controller}
			One possibility for setting a control-related compliance is the impedance control, which was introduced in \cite{Hogan.1984} and extended in \cite{AlbuSchaffer.2007, Ott.2008} for redundant, serial, torque-controlled robots. 
			A Cartesian impedance controller for PRs is chosen from \cite{Taghirad.2013} to parameterize intuitively the robot environmental dynamics on the mobile platform translationally and rotationally. Since direct drives are used without gearbox and resulting friction, joint torque control via the motor current is permissible. 
			The actuation forces in platform coordinates are given by
			\begin{equation} \label{eq_ImpRegX}
				\boldsymbol{F}_\mathrm{m} = \hat{\boldsymbol{c}}_x + \hat{\boldsymbol{g}}_x + \hat{\boldsymbol{M}}_x \ddot{\boldsymbol{x}}_\mathrm{d} + \hat{\boldsymbol{F}}_{\mathrm{fr},x} + \boldsymbol{K}_\mathrm{d} \boldsymbol{e}_x + 	\boldsymbol{D}_\mathrm{d} \dot{\boldsymbol{e}}_x 
			\end{equation}
			with the compensation of the dynamics components and the control deviation $\boldsymbol{e}_x{=}\boldsymbol{x}_\mathrm{d}{-}\boldsymbol{x}$ consisting of the setpoint and actual pose. 
			The drive torques demanded by the control can be calculated via $\boldsymbol{\tau}_\mathrm{a}{=}\boldsymbol{J}_{x,q_\mathrm{a}}^\mathrm{T}\boldsymbol{F}_\mathrm{m}$. 
			The desired stiffness matrix ${\boldsymbol{K}_\mathrm{d}{=}\mathrm{diag}(k_{\mathrm{d},1},k_{\mathrm{d},2}, \dots ,k_{\mathrm{d},m}){>}\boldsymbol{0}}$ is used together with the inertia matrix for the factorization damping design \cite{AlbuSchaffer.2003}
			\begin{align} 
				\boldsymbol{D}_\mathrm{d} = \tilde{\boldsymbol{M}}_{x} \boldsymbol{D}_\xi \tilde{\boldsymbol{K}}_\mathrm{d} + 	\tilde{\boldsymbol{K}}_\mathrm{d} \boldsymbol{D}_\xi \tilde{\boldsymbol{M}}_{x},
			\end{align} 
			with $\boldsymbol{M}_x{=}\tilde{\boldsymbol{M}}_{x} \tilde{\boldsymbol{M}}_{x}$ and $\boldsymbol{K}_\mathrm{d}{=}\tilde{\boldsymbol{K}}_\mathrm{d} \tilde{\boldsymbol{K}}_\mathrm{d}$.
			By designing $\boldsymbol{D}_\xi{=}\mathrm{diag}\{D_{\xi,i}\}$, the desired modal damping behavior can thus be obtained. 
			By assuming well-identified dynamics in (\ref{eq_ImpRegX}), the closed-loop error dynamics results as
			\begin{align}
				{\boldsymbol{M}_x ( \ddot{\boldsymbol{x}} {-}\ddot{\boldsymbol{x}}_\mathrm{d} ) {+} \boldsymbol{D}_\mathrm{d} ( \dot{\boldsymbol{x}} {-}\dot{\boldsymbol{x}}_\mathrm{d} ) {+} \boldsymbol{K}_\mathrm{d} ( \boldsymbol{x} {-} \boldsymbol{x}_\mathrm{d} ) {=} \boldsymbol{F}_\mathrm{ext}.}
			\end{align}
			This corresponds to a system with the external force as input and the control deviation as output. 
		\subsection{Generalized-Momentum Observer} \label{sec:observer}
			The initial point of the observer, introduced by De Luca \cite{Luca.2003}, is a residual of the generalized momentum $\boldsymbol{p}_x{=}\boldsymbol{M}_x \dot{\boldsymbol{x}}$ which is set up in the operational space coordinate $\boldsymbol{x}$ as the minimal coordinate for the dynamics of fully-parallel robots.
			The derivative of the residual w.r.t. time is $\dot{\hat{\boldsymbol{F}}}_\mathrm{ext} {=} \boldsymbol{K}_{\mathrm{o}} (\dot{\boldsymbol{p}}_x {-} \dot{\hat{\boldsymbol{p}}}_x )$ with $\boldsymbol{K}_\mathrm{o}{=}\mathrm{diag}(k_{\mathrm{o},1},k_{\mathrm{o},2}, \dots, k_{\mathrm{o},m}){>}\boldsymbol{0}$ chosen as observer gain matrix \cite{Luca.2003}. 
			Transforming (\ref{eq_dyn}) to $\hat{\boldsymbol{M}}_x\ddot{\boldsymbol{x}}$ and substituting it into the integral of $\dot{\hat{\boldsymbol{F}}}_\mathrm{ext}$ over time, leads to 
			\begin{align} 
				\hat{\boldsymbol{F}}_\mathrm{ext} &{=} \boldsymbol{K}_\mathrm{o} ( \hat{\boldsymbol{M}}_x \dot{\boldsymbol{x}} {-} 	\int_{0}^t \boldsymbol{F}_\mathrm{m} {-} \hat{\boldsymbol{\beta}} + \hat{\boldsymbol{F}}_\mathrm{ext} \mathrm{d}\tilde{t} ), \\ 
				\nonumber
				\hat{\boldsymbol{\beta}} &{=} \hat{\boldsymbol{g}}_x {+} \hat{\boldsymbol{F}}_{\mathrm{fr},x} {+}( \hat{\boldsymbol{C}}_x {-}\dot{\hat{\boldsymbol{M}}}_x )\dot{\boldsymbol{x}} {=} \hat{\boldsymbol{g}}_x {+} 	\hat{\boldsymbol{F}}_{\mathrm{fr},x}{-}\hat{\boldsymbol{C}}_x^\mathrm{T} \dot{\boldsymbol{x}}
			\end{align}
			with $\dot{\hat{\boldsymbol{M}}}_x{=}\hat{\boldsymbol{C}}_x^\mathrm{T}{+}\hat{\boldsymbol{C}}_x$ \cite{Haddadin.2017, Ott.2008}. Under the condition $\hat{\boldsymbol{\beta}}{\approx}\boldsymbol{\beta}$ it follows 
			\begin{align}
				\boldsymbol{K}_\mathrm{o}^{-1} \dot{\hat{\boldsymbol{F}}}_\mathrm{ext} + 	\hat{\boldsymbol{F}}_\mathrm{ext}=\boldsymbol{F}_\mathrm{ext}, 
			\end{align}
			which corresponds to a linear and decoupled error dynamics of the generalized-momentum observer (MO), exponentially converging to the external force projected to platform coordinates. 
		\subsection{Kalman Filter} \label{sec:observer_kf}
			To account for modeling errors and measurement noise, a Kalman filter (KF) with the state-space model
			\begin{align} \label{eq_kalman_state_equation}
				\left[
					\begin{array}{c} 
						\dot{\boldsymbol{p}}_x \\ 
						\dot{\hat{\boldsymbol{F}}}_\mathrm{ext} \\ 
					\end{array}
				\right] = 
				\left[
					\begin{array}{cc}
						\boldsymbol{0} & \boldsymbol{I}\\
						\boldsymbol{0} & \boldsymbol{0}\\
					\end{array}
				\right] 
				\left[
						\begin{array}{c} 
						\boldsymbol{p}_x \\ 
						\hat{\boldsymbol{F}}_\mathrm{ext} \\ 
					\end{array}
				\right] + 
				\left[
					\begin{array}{c}
						\boldsymbol{F}_\mathrm{m} - \hat{\boldsymbol{\beta}}\\				
						\boldsymbol{0}\\
					\end{array}
				\right],
			\end{align}
			 the covariance matrices $\boldsymbol{Q},\boldsymbol{R}{\in} \mathbb{R}^{2m \times 2m}$ of the process and the measurement noise and the output vector $\boldsymbol{y}{=}\boldsymbol{p}_x$ is taken from \cite{Wahrburg.2015} and adapted in the operational space. Here, the force estimation is done only by the update step in the algorithm of the Kalman filter. The system is observable according to Kalman's observability criterion, since the observability matrix consisting of the constant state and output matrix always has full rank. 	
		\subsection{Sliding-Mode Momentum Observer}	\label{sec:observer_sosml}
			To obtain a finite-time behavior instead of the exponential convergence of the classical generalized-momentum observer, a second-order sliding-mode with linear terms (SOSML) is taken from \cite{Garofalo.2019}. 
			The equation for observation is described by 
			\begin{align}
				%
				\dot{\hat{\boldsymbol{p}}}_x &{=} 		( \boldsymbol{F}_\mathrm{m} {-} \hat{\boldsymbol{\beta}} ) {-} \boldsymbol{T}_1 	|\tilde{\boldsymbol{p}}_x|^{\frac{1}{2}} \mathrm{sgn} (\tilde{\boldsymbol{p}}_x) {-} \boldsymbol{T}_2 \tilde{\boldsymbol{p}}_x {+} \hat{\boldsymbol{F}}_\mathrm{ext},\nonumber \\
				\label{eq_SOSML}
				\dot{\hat{\boldsymbol{F}}}_\mathrm{ext} &{=} {-}\boldsymbol{S}_1 \mathrm{sgn} (\tilde{\boldsymbol{p}}_x) {-} 	\boldsymbol{S}_2 \tilde{\boldsymbol{p}}_x
			\end{align}	
			with the positive diagonal matrices $\boldsymbol{S}_i, \boldsymbol{T}_i{\in} \mathbb{R}^{m \times m}$ and ${\tilde{\boldsymbol{p}}_x {=} \hat{\boldsymbol{p}}_x {-} \boldsymbol{p}_x}$. Provided that the perturbation terms in equation (\ref{eq_SOSML}) are globally bounded, the matrices $\boldsymbol{S}_i,\boldsymbol{T}_i$ can be designed according to the inequalities in \cite{Moreno.2008,Garofalo.2019} to achieve global finite-time convergence to the equilibrium point $[\begin{array}{cc} \tilde{\boldsymbol{p}}_x^\mathrm{T} & \hat{\boldsymbol{F}}_\mathrm{ext}^\mathrm{T}{-}\boldsymbol{F}_\mathrm{ext}^\mathrm{T} \end{array}]^\mathrm{T}{=}\boldsymbol{0}$.
	\section{Experimental Validation} \label{section_validation}
		Starting with the presentation of the test bench (\ref{ssec:ExpSetup}), the results of the identification of the motor torque constant are described (\ref{ssec:IdentiMotTorCon}). 
		Afterward, the disturbance observation and impedance control on contact with the platform and then on the leg chains are shown (\ref{ssec:ImpControlDistObs}). 
		Finally, the results of the collision and clamping detection are discussed (\ref{ssec:CollClampDet}).
		\subsection{Experimental Setup} \label{ssec:ExpSetup}
			The active joints of the 3-\underline{R}RR PR are actuated by three torque motors\footnote{KTY6288.4 from Georgii Kobold} (gearless synchronous motors). 
			The angular positions are measured by absolute encoders\footnote{ECN1313 from Dr. Johannes Heidenhain} with a system accuracy of $\SI{0.0056}{\degree}$ and are integrated into the data communication of the servo drive\footnote{S600 from Kollmorgen Europe}, which are then numerically differentiated and low-pass filtered with $\SI{30}{\hertz}$ for the velocity computation. 
			Passive joint angles are measured via incremental encoders\footnote{RI36-H from Hengstler} with an accuracy of $\SI{0.1}{\degree}$, which are integrated into the data communication by a channel encoder interface\footnote{EL5101 from Beckhoff Automation} with 16 bits. 
			Two force-torque sensors\footnote{KMS40 from Weiss Robotics}$^{,}$\footnote{Mini40 from ATI Industrial Automation} (FTS) are used to measure the contact forces by $\SI{500}{\hertz}$ and $\SI{1000}{\hertz}$ to validate the observers and the impedance control. 
			Their measurement ranges are $\pm\SI{120}{\newton}/\pm\SI{3}{Nm}$ and $\pm\SI{240}{\newton}/\pm\SI{4}{Nm}$. 
			The communication architecture is based on the EtherCAT protocol and the open-source tool EtherLab\footnote{\url{https://www.etherlab.org}} which was modified with an external-mode patch and a shared-memory real-time interface\footnote{\url{https://github.com/SchapplM/etherlab-examples}}. 
			Thus, a ROS package\footnote{\url{https://github.com/ipa320/weiss_kms40}} for the FTS can be integrated into the communication with the control system in \textsc{Matlab}/Simulink.
			\begin{figure}[tb!]
			\vspace{1.5mm} 
			\centering
			\includegraphics[width=\columnwidth]{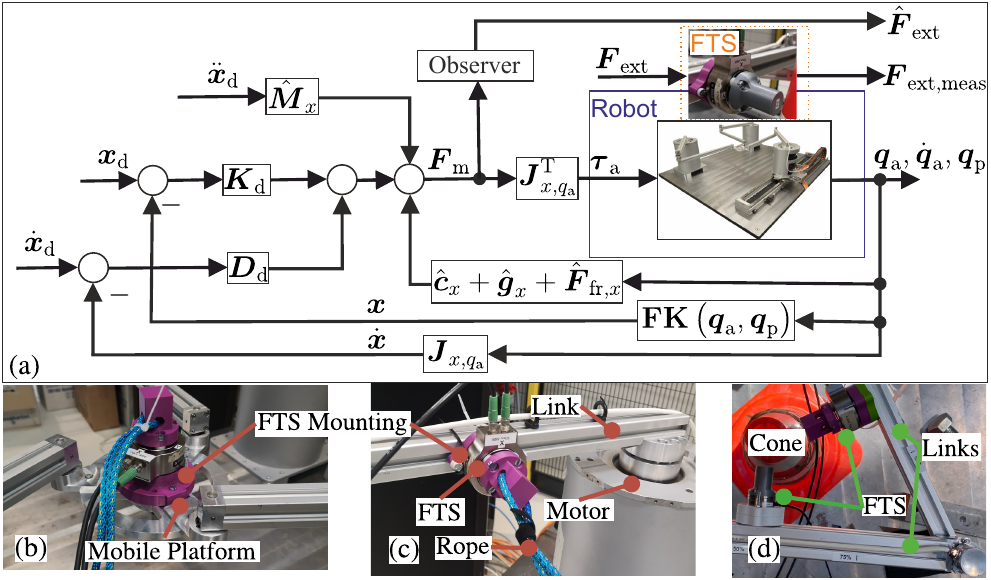}
			\caption{(a) Block diagram with the experimental setup and pulling a rope on the platform (b), on a link (c) and clamping test (d)}
			\label{fig_Ablauf_3PRRR}
			\vspace{-1.75mm}
			\end{figure}	
			Figure~\ref{fig_Ablauf_3PRRR}(a) represents the block diagram of the Cartesian impedance control and the observer. The error is calculated at a sampling rate of $\SI{1}{\kilo\hertz}$ in the operational space and forms the input to the multi-axis control. 
			The parameterization of the Cartesian impedance controller is set to $\boldsymbol{K}_\mathrm{d}{=}\mathrm{diag}(\SI{2}{\newton/ \milli\meter}, \SI{2}{\newton/\milli\meter}, \SI{85}{Nm/\radian})$ and a critical damping of $\boldsymbol{D}_\xi{=}\mathrm{diag}(1,1,1)$. 
			The MO is parameterized with $k_{\mathrm{o},i}{=}\frac{1}{\SI{50}{\milli \second}}$.
			For the following experiments, a rope is pulled manually, which is tied to the FTS on the mobile platform and subsequently at the robot links, like shown in Fig.~\ref{fig_Ablauf_3PRRR}(b)--(c). 
			The coordinate system of the FTS is rotated into the inertial coordinate system via the measured $\boldsymbol{q}$ or calculated $\boldsymbol{x}$, depending on the mounting position. 
			To validate the disturbance observer and the specified impedance, the motor torque constant is identified at the beginning. For this purpose, the FTS is mounted at each link and the measured forces are transformed using (\ref{eq_trafo_link_mP_Drives}) to the platform and drives.
		\subsection{Identification of the Motor Torque Constant} \label{ssec:IdentiMotTorCon}
			\begin{figure}[t!]
			\vspace{2.5mm}
			\centering
			\subfloat[
			Transformed and linearly fitted torques of the third axis over the current 
			\label{fig_torque_over_current}
			]
			{
				\includegraphics[width=0.95\columnwidth]{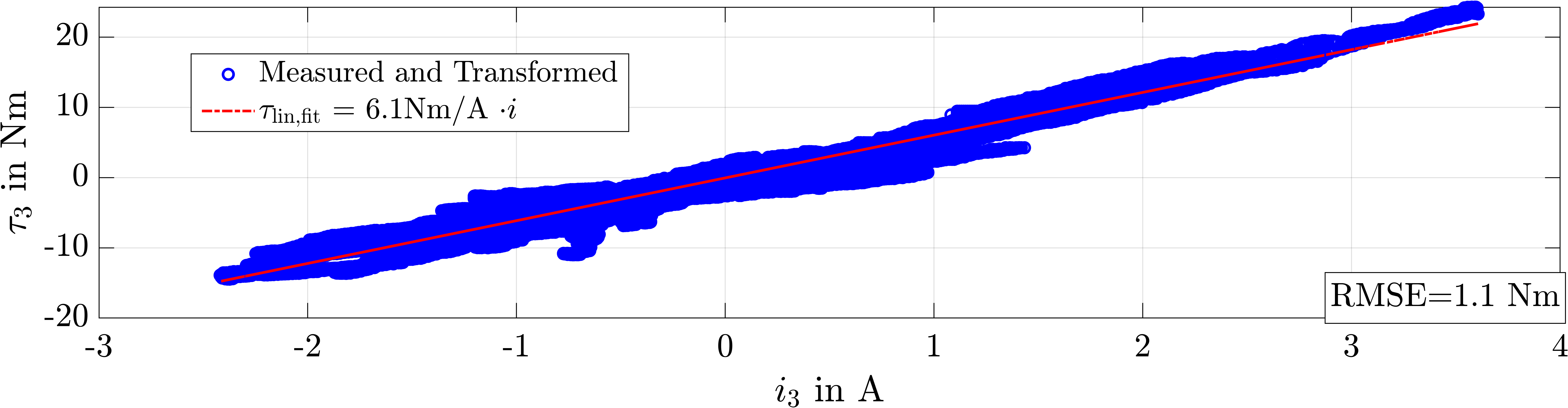} 
			}
			\hspace{-1em}
			\subfloat[
			 Box plot of the error between estimated force $\hat{f}_x$ by the MO and measured force $f_x$ via the FTS over the affected body (Abbreviations are: C$i$L$j$ is the $j$-th link of the $i$-th chain and MP is the mobile platform)
			\label{fig_forceEstimationvsMeasurement}
			]
			{
				\includegraphics[width=0.95\columnwidth]{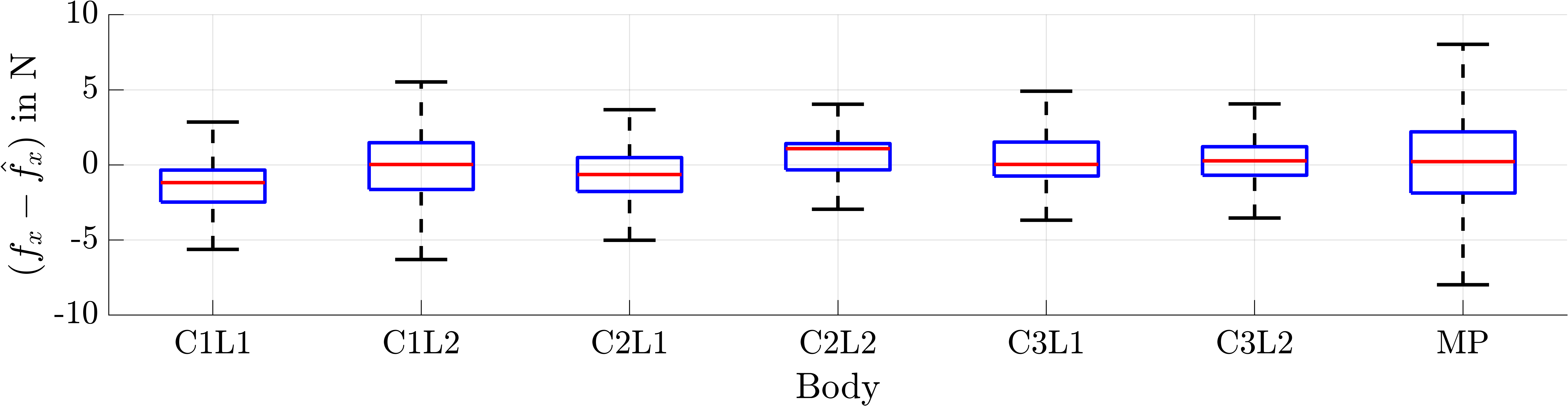} 
			} 
			\caption{Identification in (a) and validation in (b) of the motor torque constant in two different configurations}
			\label{fig_torque_over_current_forceEstimationvsMeasurement}
			\vspace{-2mm}
			\end{figure}
			In Fig.~\ref{fig_torque_over_current_forceEstimationvsMeasurement} the results of the identification and validation of the motor torque constant of the third drive are shown. 
			The identification in Fig.~\ref{fig_torque_over_current_forceEstimationvsMeasurement}(a) is based on the contact forces transformed to the drives at the six links and the platform. 
			Linear regression is used to determine a motor torque constant of $\SI{6.1}{\newton\meter / \ampere}$, whose RMSE is $\SI{1.1}{\newton\meter}$. 
			Validation is performed in a different configuration, where contacts are also applied on all seven bodies and the errors $(f_x{-}\hat{f}_x)$ are shown over the seven bodies in Fig.~\ref{fig_torque_over_current_forceEstimationvsMeasurement}(b). 
			The results show that the absolute maximum error is less than $\SI{10}{\newton}$. 
			Possible reasons for the errors are modeling inaccuracies of the dynamics, such as cogging torques or assuming the same inertia of the three leg chains even though the FTS is mounted.
		\subsection{Impedance Control and Disturbance Observer} \label{ssec:ImpControlDistObs}
			\begin{figure}[b!]
			\centering
			\subfloat[
						Measured forces $f_y$ over the position $r_y$ with predefined and fitted stiffnesses $K_\mathrm{d},K_\mathrm{fit}$ from 0.1--$\SI{2}{\newton /\milli \meter}$ on the mobile platform 
						\label{fig_LowImpedances}
					]
					{
						\includegraphics[width=0.95\columnwidth]{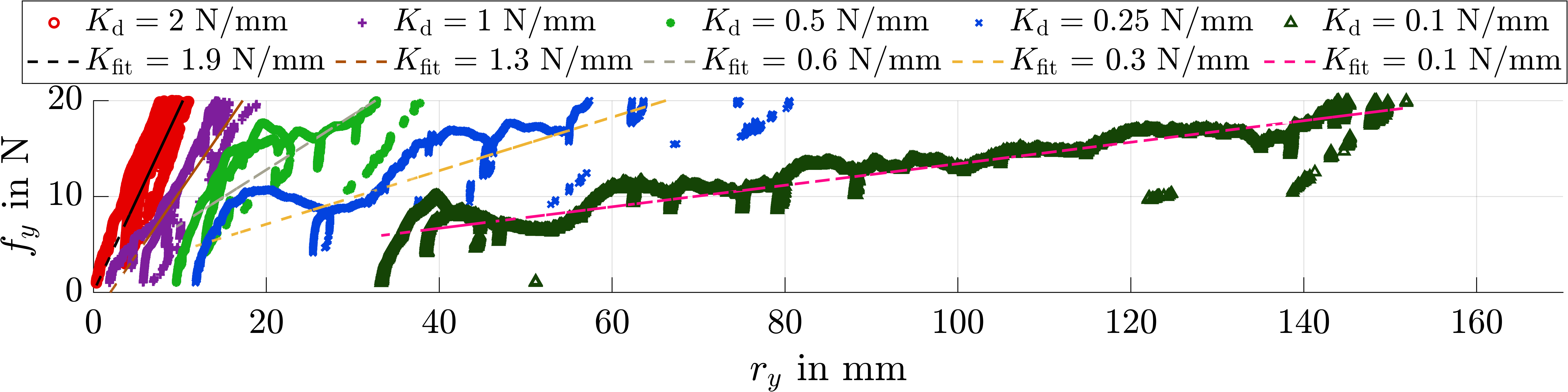} 
					}
			\hspace{-1em}
			\subfloat[
						The estimation error $(f_y-\hat{f}_y)$ over the stiffness $K_\mathrm{d}$
						\label{fig_DistObserver}
					]
					{
						\includegraphics[width=0.95\columnwidth]{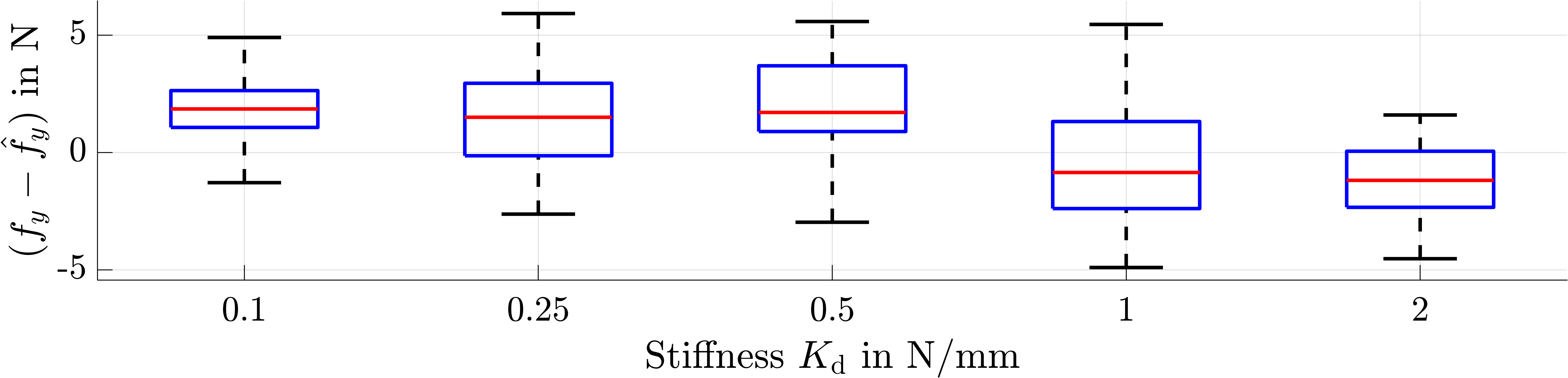} 
					}
			\caption{Results of the low controller impedances}
			\label{fig_LowImpedances_DistObserver}
			\end{figure}
			\begin{figure*}[t!] 
			\centering
			\subfloat[
			Collision at the platform
			\label{fig_CollDet_results_mobPlat}
			]
			{
				\includegraphics[width=\columnwidth]{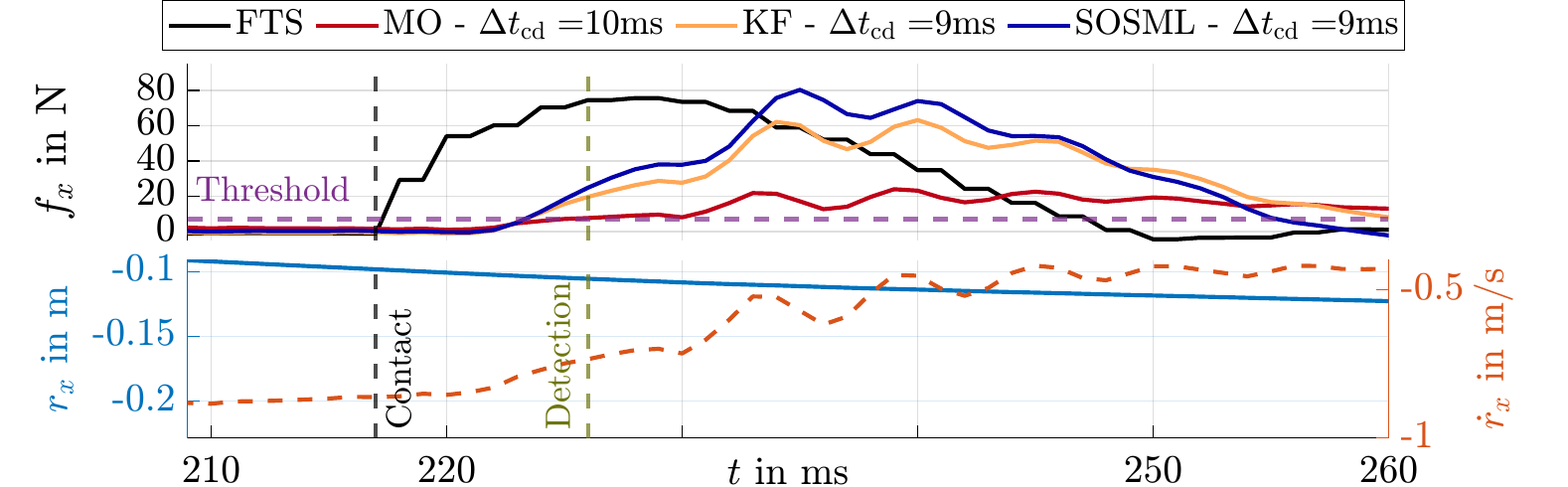} 
			}
			\hspace{-1em}
			\subfloat[
			Collision at the first link
			\label{fig_CollDet_results_C2L1}
			]
			{
				\includegraphics[width=\columnwidth]{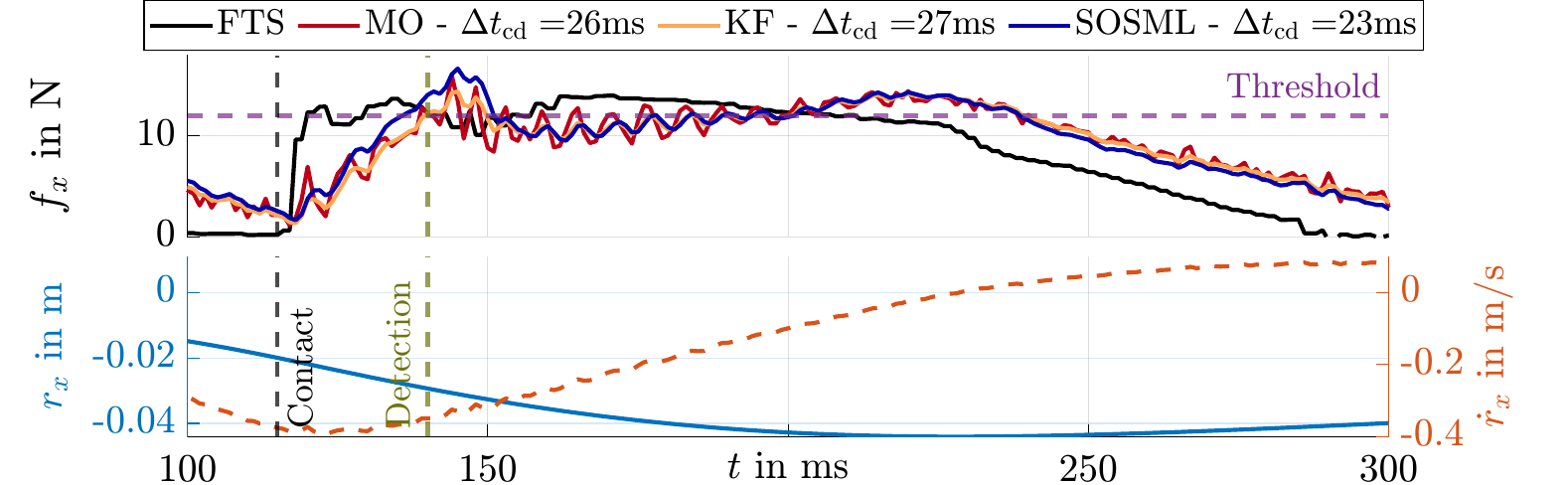} 
			}
			\hspace{-1em}
			\subfloat[
			Collision at the second link
			\label{fig_CollDet_results_C2L2}
			]
			{
				\includegraphics[width=\columnwidth]{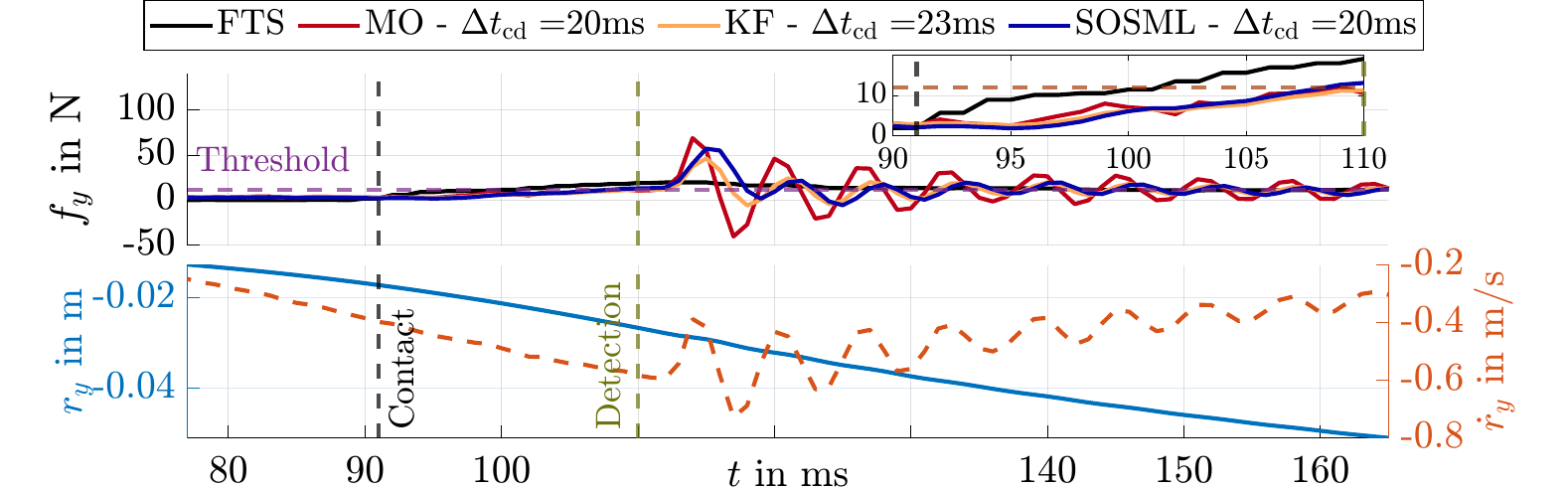} 
			}
			\hspace{-1em}
			\subfloat[
			Clamping between links
			\label{fig_ClampDet_results}
			]
			{
				\includegraphics[width=\columnwidth]{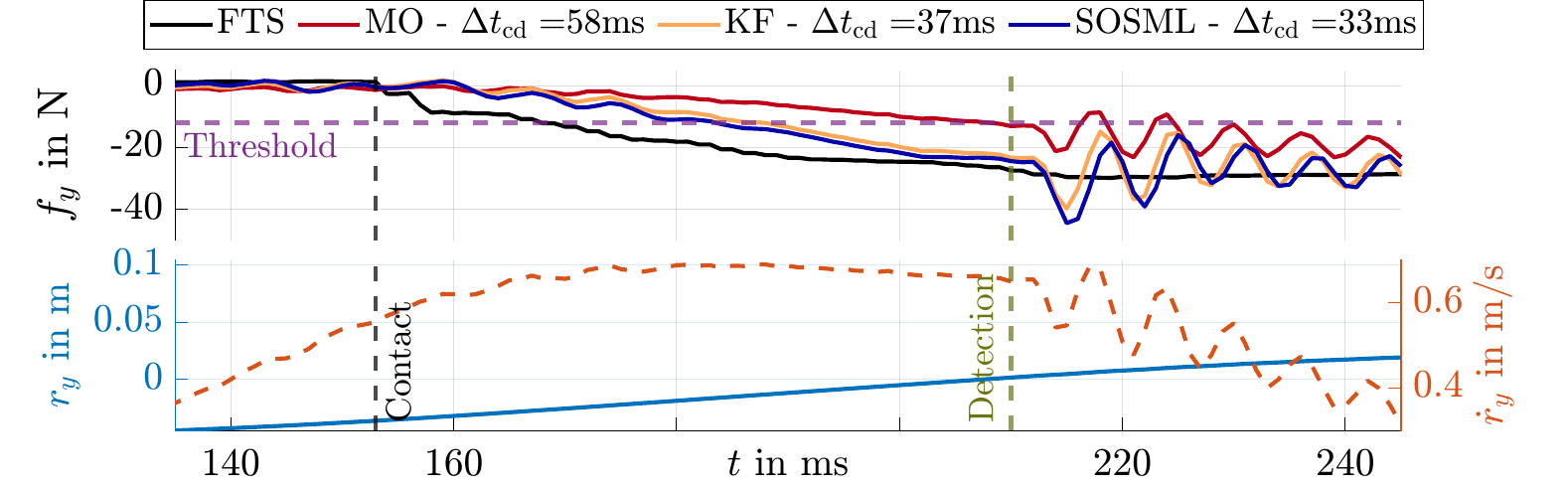} 
			}
			\caption{Estimated and measured forces $\hat{f},f$, positions $r$ and velocities $\dot{r}$ of the mobile platform for evaluating the contact detection. As soon as the observed force by the MO exceeds the threshold (purple dashed line), the robot movement is stopped. For comparison of the observers, the KF ($Q_{f}{=}10$) and the SOSML are presented. $\Delta t_\mathrm{cd}$ is the time difference between contact occurrence (black dashed) and detection by the corresponding observer}
			\label{fig_CollDet_results}
			\vspace{-1.5mm}
		\end{figure*}
			In the following, the measurements of stepwise constant forces on the mobile platform are used to validate the stiffness $\boldsymbol{K}_\mathrm{d}$ of the impedance control. 
			Figure~\ref{fig_LowImpedances_DistObserver}(a) depicts the evolution of the measured force $f_y$ of the FTS versus the position $r_y$ for different controller stiffnesses from $\SI{0.1}{}$ to $\SI{2}{\newton /\milli \meter}$. 
			A linear regression is performed from these measurements to calculate $K_\mathrm{fit}$ for each selected impedance. 
			The stiffness is particularly visible in the low force ranges. 
			The errors of the MO compared to the FTS are shown over the stiffnesses in Fig.~\ref{fig_LowImpedances_DistObserver}(b), where small differences of up to $\SI{5}{\newton}$ can be seen. 
			Compared to robots with gearboxes, friction effects occur to a lesser extent here due to the direct drive, so more accurate conclusions can be drawn from the motor current to the joint torque on the link side.
		\subsection{Collision and Clamping Detection} \label{ssec:CollClampDet}
			In this section results from collisions and clamping with a compliant traffic cone during the robot movement are considered for the evaluation of a contact detection. 
			Due to the deviations between observed and measured external force of the previous results, thresholds for contact detection are defined empirically by $\boldsymbol{\epsilon}_\mathrm{ext}^\mathrm{T}{=}[\SI{12}{\newton},\SI{12}{\newton},\SI{1}{Nm}]$. 
			As soon as $|\hat{F}_{\mathrm{ext},i}|{>}\epsilon_{\mathrm{ext},i}$, the controller's desired torques are set equal to zero corresponding to a reaction strategy of gravity compensation for robots under influence of gravity. 
			\begin{figure}[b!]
				\centering
				\includegraphics[width=\columnwidth]{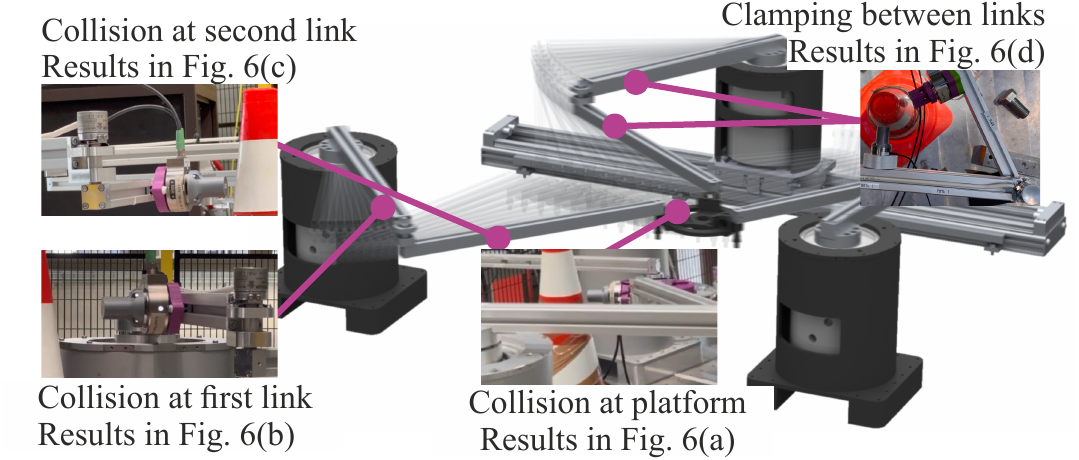}
				\caption{Contact forces on the mobile platform, first and second link to test collision and clamping detection by the observer in robot motion}
				\label{fig_CollDet_ablauf}
			\end{figure}
		
			Detection and reaction are tested on the mobile platform and links as shown in Fig.~\ref{fig_CollDet_ablauf}. 
			The results of the observed and measured forces at the different locations, as well as positions and velocities of the platform are shown in Fig.~\ref{fig_CollDet_results}. 
			The times of collision occurrence and detection are also shown, as well as their time delays $\Delta t_\mathrm{cd}$. 
			In all cases, contact is detected with a time delay of $\SI{9}{}$--$\SI{58}{\milli \second}$, after which the execution of the movement is interrupted. 
			In Fig.~\ref{fig_CollDet_results}(a), the platform reaches a maximum velocity of up to $\SI{0.9}{ \meter / \second}$ until the collision occurs at $t{\approx} \SI{217}{\milli \second}$. 
			The estimates of all three observers almost simultaneously exceed the threshold within $\SI{10}{\milli \second}$, so that a maximum force of $\SI{80}{\newton}$ is measured by the FTS (black line) due to the motion termination. 
			The results of the structure collisions are shown in Fig.~\ref{fig_CollDet_results}(b)--(c). 
			It can be seen that the approach with the SOSML detects the contact the fastest, but the forces estimated by the observers show a similar pattern, so that the contacts are detected within $20$--$\SI{27}{\milli \second}$. 
			A difference regarding $\Delta t_\mathrm{cd}$ is in the results of a structural clamping between two links in Fig.~\ref{fig_CollDet_results}(d). 
			The approach with the SOSML leads to a motion stop after $\Delta t_\mathrm{cd}{=}\SI{33}{\milli\second}$ compared to the MO with $\Delta t_\mathrm{cd}{=}\SI{58}{\milli\second}$. 
			It is noticeable that oscillations in the estimated forces occur after the detection $t{=}\SI{216}{\milli \second}$. 
			Oscillations are strongest in Fig.~\ref{fig_CollDet_results}(c) starting at $t{=}\SI{110}{\milli \second}$. 
			These are due to the velocities shown in the lower plot, so that the momentum also oscillates. 
	\section{Conclusions} \label{section_conlusions}
		This work aims to present a first step towards human-robot collaboration with parallel robots. 
		A kinetostatic model of arbitrary points on the robot structure is designed to project the contact forces measured by force-torque sensors to the platform and (actuated) joint coordinates. 
		The experimental results show that lower Cartesian control impedances and force estimation for low contact forces based on direct drives and the motor current are feasible with deviations up to $\SI{5}{\newton}$, complying with HRC requirements. 
		Contact detection and motion stopping are successful and the time delays between contact occurrence and detection are in the range $9$--$\SI{58}{\milli \second}$ at a maximum velocity of $\SI{0.9}{\meter/\second}$, in which the robot reacts to the object in an impedance-controlled manner. 
		Further reaction strategies are being envisaged that may require information about the contact location to perform a retraction movement in the occurrence of a collision or the opening of leg chains in the event of a clamp. 
	
	\addtolength{\textheight}{-0cm}   

	\section*{ACKNOWLEDGMENT}
	The authors acknowledge the support by the German Research Foundation (Deutsche Forschungsgemeinschaft) under grant number 444769341.
	
	\bibliographystyle{IEEEtran}
	\bibliography{literatur}

\end{document}